# On-Device Transfer Learning for Personalising Psychological Stress Modelling Using a Convolutional Neural Network

Kieran Woodward[1] Eiman Kanjo[1] David J. Brown[1] T.M. McGinnity[1]

## Abstract

Stress is a growing concern in modern society adversely impacting the wider population more than ever before. The accurate inference of stress may result in the possibility for personalised interventions. However, individual differences between people limits the generalisability of machine learning models to infer emotions as people's physiology when experiencing the same emotions widely varies. In addition, it is time consuming and extremely challenging to collect large datasets of individuals' emotions as it relies on users labelling sensor data in real-time for extended periods. We propose the development of a personalised, cross-domain 1D CNN by utilising transfer learning from an initial base model trained using data from 20 participants completing a controlled stressor experiment. By utilising physiological sensors (HR, HRV EDA) embedded within edge computing interfaces that additionally contain a labelling technique, it is possible to collect a small real-world personal dataset that can be used for on-device transfer learning to improve model personalisation and cross-domain performance.

## 1 Introduction

Modern lifestyles are contributing heavily to increased levels of daily stress with more adults experiencing work-related stress (Perkbox, 2018) and more students experiencing poor mental health (The Physiological Society, 2017) than ever before making it more vital than ever to manage such pressures. The ability to infer emotions is an exciting proposition as it could enable better management of wellbeing by providing real-time interventions. Electrodermal Activity (EDA) is often used to infer stress as it directly correlates to the sympathetic nervous system (Schumm et al., 2010), along with Heart Rate Variability (HRV). The latter represents the variation in time between heartbeats and when HRV is reduced the user is more likely to be stressed (Wijsman et al., 2011). Non-invasive physiological sensors that can easily be embedded within wearables or edge computing interfaces including those that measure Heart Rate (HR), HRV and EDA present a significant opportunity to be widely adopted unlike other invasive approaches using Electrocardiography (ECG) and Electroencephalogram (EEG) which cannot be widely adopted outside of controlled experiments.

Sensor streams paired with sufficiently trained Artificial Intelligence (AI) models may make the possibility of ubiquitous emotion inference a reality. However, labelled emotion data can be challenging to collect as unlike images which can be crowdsourced (Google, 2019) emotion data must be collected and labelled in real-time. Longitudinal data collection poses even greater challenges as it relies on multiple users continually recording their emotions while simultaneously wearing sensors for extended periods. These drawbacks result in limited labelled data which significantly impacts model accuracy as deep learning requires a vast labelled training dataset. Furthermore, emotions are always personal and individualistic with individual users experiencing large variation in physiological parameters when experiencing the same emotions thus making a generalised model additionally challenging.

Recent advancements in wearable EDA and HRV sensors have enabled new research using machine learning classifiers to infer stress (Healey & Picard, 2005). However, this does not consider the personalisation of models or the possibility of the models being used in different domains for example a model developed using a controlled experiment dataset may not perform as accurately when used in real-world environments. It is essential to develop subjectdependent models for stress detection as when working with a heterogeneous population there are numerous physiological factors such as age, gender and diet that result in a variation of physiological data (Picard et al., 2001).

Transfer learning has most commonly been used for object recognition (Oquab et al., 2014), human activity recognition (Sargano et al., 2017) and speech recognition (Wang & Zheng, 2016). However it may also be used to address the challenges of domain adaption and personalisation by using a pre-trained model from a different domain and transferring the learned knowledge to the target domain. Caruana (Caruana, 1997) introduced multi-task learning that uses domain information contained in the training signals of related tasks. Convolutional Neural Networks (CNN) are commonly used in transfer learning approaches where they are initially trained on a vast dataset and then the last fully-connected layer is removed and further trained on the smaller target dataset. A pre-trained base model removes the requirement for a large target dataset while simultaneously decreasing the time required to train the model.

The possibility for transfer learning to be used to personalise affective models has previously been explored. However most work has focused on personalising EEG signals, where transfer learning approaches have improved model accuracy by 19% (Zheng & Lu) and 12.72% (Li et al., 2019) while also reducing the amount of data required to train the models. Transfer learning can be used to help alleviate scarce data as by using decision trees, data from similar subjects can be used to improve accuracy by around 10% although if data from dissimilar subjects is used it can have a negative impact on the model accuracy (Maxhuni et al., 2016). To ensure negative transfer learning that degrades the performance of the model does not occur, a conditional transfer learning framework has been developed that assesses individual's transferability against all other individual's data within the dataset. When tested the conditional transfer learning model identified up to 16 individuals who could benefit from an original dataset of 18 individuals' EEG signatures, improving classification accuracy by around 15% (Lin & Jung, 2017).

With the rise of edge computing it is now possible to train models using transfer learning and fine-tune models ondevice. iSelf (Sun et al., 2015) applied a local transfer learning approach within a smartphone application to infer emotions by learning smartphone usage habits. Users provided only a few sample labels and when utilising transfer learning the model achieved 75% accuracy. Applying a transfer learning approach on-device greatly improves privacy as no data is externally transmitted which is vital when collecting personal wellbeing data.

To achieve a personalised model the development of an initial base model to infer emotion trained using HRV, HR and EDA data from a controlled experiment is proposed, followed by an on-device transfer learning approach to personalise the model using real-world labelled data collected from individuals using edge computing interfaces.

## 2 EXPERIMENTAL SETUP

### 2.1 Controlled Stressor Experiment

Model adaption is a personalisation and cross-domain transfer learning problem. Here, a model trained on a dataset collected from a controlled experiment is adapted to perform the same task in a different situation where real-world physiological data will be collected from target participants.

The controlled experiment used the Montreal stress test (Dedovic et al., 2005) to induce stress in 20 participants. Participants wore HR, HRV and EDA sensors on their hand each sampled at 30Hz to collect physiological data when relaxing and experiencing stress. All participants were initially briefed before completing a 3-minute rest period, followed by 3 minutes of training how to use the system to answer maths questions before another 3-minute rest period. Participants then completed the stressor experiment where the time limit for each question was the average time taken during training reduced by 10%. The time pressure along with a progress bar showing their progress compared with an artificially inflated average were both designed to induce stress during the 10-minute experiment. Finally, participants completed a final 3-minute rest session to ensure sufficient stressed and relaxed data was collected.

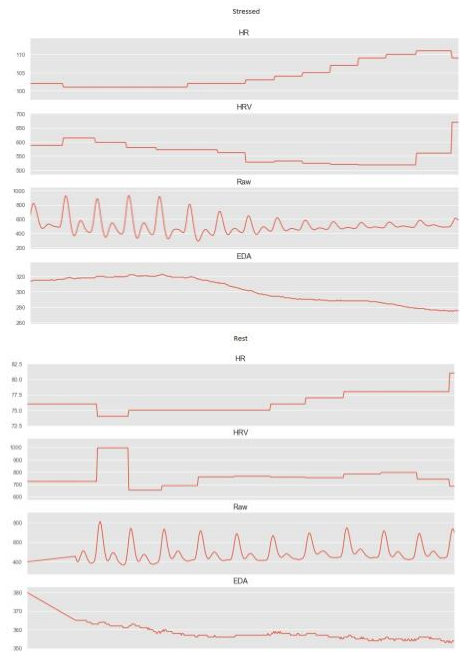

*Figure 1.* Comparison of stressed and relaxed data collected

A similar number of samples was collected of both relaxed and stressed data with 420,000 records collected of stressed data and 400,000 records of relaxed data helping to reduce bias in the classification model. Figure 1 shows a sample of 400 records representing around 13 seconds of relaxed and stressed data during the experiment. While time-series data is traditionally challenging to classify by sight there are clear trends displayed such as higher heart rate (stressed average81.2, SD-12.4; relaxed average-76.8, SD-10.7) and lower EDA (stressed average-317.2, SD-148.3; relaxed average351.9, SD-152.8). However, this was not true of all 20 participants with some experiencing more pronounced physiological changes than others indicating the need for personalised models.

## 2.2 On-device model personalisation

While a transfer learning approach can assist personalising models and improve models' performance across multiple domains, it has traditionally relied on collecting the target domain data in advance. The training is then completed externally before the model can be exported to devices.

We have devised a new approach where the target user provides labelled samples over a short period and transfer learning is then used to personalise the model on-device. Three participants were provided with physical prototypes containing the same HR, HRV and EDA sensors as used in the controlled experiment as well as two buttons to enable the real-time self-labelling of stressed and relaxed emotions. For processing the device utilised a Raspberry Pi 3, with 1GB of RAM and a 1.2GHz quad core CPU, as it is compact so can be embedded within the prototypes, sufficiently powerful to perform transfer learning and affordable.

Transfer learning can then be applied on-device to further train the base model with the new target user's data. This approach personalises the model and improves real-world domain accuracy, as the target data is the real-world data collected in the "wild" in comparison to the controlled experiment data used to train the base model.

## 3 NETWORK ARCHITECTURE

The classification of stress is a time-series classification task which takes the raw physiological signals as input and outputs a label (stressed or relaxed) for each sequence. Deep learning particularly CNNs present many opportunities for the classification of emotions. Traditionally CNNs have been used to classify two-dimensional data but 1 dimensional (1D) CNNs can learn from raw time-series data without feature extraction being first required. The raw input data is first divided into segments of fixed lengths with sliding windows used to avoid semantic segmentation.

The stressed and relaxed data collected from the experiment was used to train a 1D CNN over 50 epochs using 10-fold cross validation achieving 82.5% accuracy. The network architecture consists of multiple 1 dimensional convolutional layers followed by max pooling operations. A dropout layer with a rate of 0.3 is included to prevent overfitting before the softmax activation function as shown in figure 2.

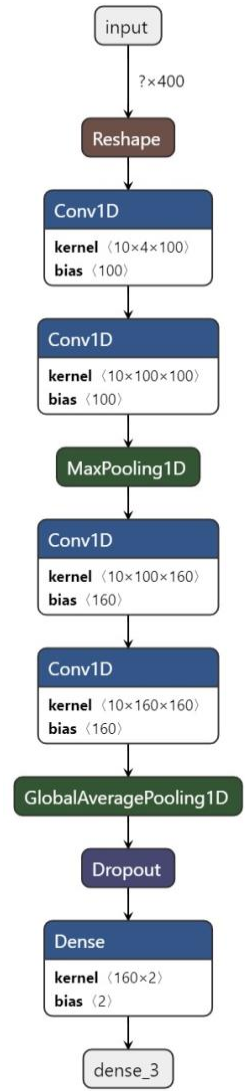

*Figure 2.* 1D CNN model architecture

During the self-labelling period each of the 3 target users labelled similar levels of data in comparison with 1 user participating in the controlled experiment. Thus, each of the 3 target users provided sufficient data to personalise the models and ensure the models function in the real-world target domain. The base model was re-trained after the data collection period where transfer learning was performed on-device using the real-world data. In order to achieve the transfer the fully connected layer in the base pre-trained model was removed from the network and two fully connected layers were added forming an adaption layer with the first having a size of 160 and the second having a size of 2; for the stressed and relaxed classes.

The on-device processing was slower than expected taking an average of 25 minutes to train the model using the transfer learning approach due to the limited specifications of the Raspberry Pi. However, as this training only needs to be completed once it is not a major limitation. This approach remains simpler than returning the devices to train the model externally and then provide users with the device embedding the personalised model.

## 4 RESULTS

The final model accuracy of the three targets were very similar with a variance of 0.1% and an average accuracy of 93.9% a significant improvement over the base model accuracy as shown in table 1. The f1-scores of the 3 models are lower and vary slightly although they consistently show high precision and recall.

*Table 1.* Accuracy of model when personalised for individual users and tested with all three target users' data

|  | USER 1 | USER 2 | USER 3 |
| --- | --- | --- | --- |
| ACCURACY | 93.9% | 93.8% | 93.9% |
| F1-SCORE | 0.91 | 0.85 | 0.9 |

Figure 3 shows the confusion matrix after transfer learning had been performed for the three target users demonstrating that few misclassification errors occurred during testing.

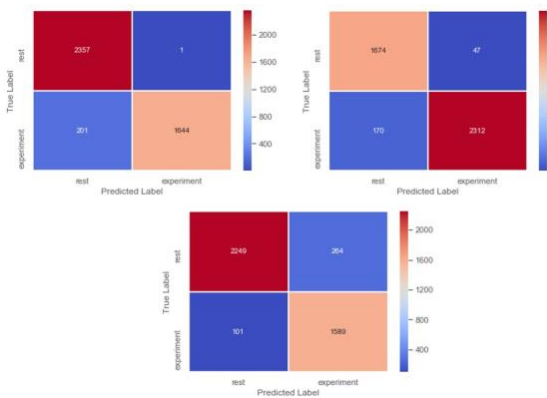

*Figure 3.* Confusion matrix for target 1, target 2 and target 3

To ensure the transfer learning approach had successfully learned the new domain the original base model trained with the controlled experiment data was tested with the three target users' real-world data. The accuracy achieved for the three users was 55% with an f1-score of 0.58, 52% with an f1-score of 0.53 and 40% with an f1-score of 0.41. These results show the transfer learning approach used has greatly improved the classification accuracy of the target real-world domain.

The aim of using transfer learning was to additionally personalise the model. To demonstrate the model personalisation each target user's model was tested with all target users' data as shown in 2. The results confirm a transfer learning approach has developed a cross-domain personalised model as there is a significant accuracy improvement when the personalised model is tested using the same user's data.

*Table 2.* Accuracy of model when personalised for individual users and tested with all three target users' data

|  | MODEL 1 | MODEL 2 | MODEL 3 |
| --- | --- | --- | --- |
| USER 1 DATA | 93.9% | 67% | 70% |
| USER 2 DATA | 54% | 93.8% | 63% |
| USER 3 DATA | 48% | 88% | 93.9% |

Overall a transfer learning approach has shown to improve model performance, demonstrating its ability to personalise the affective model and work across domains. The 11% accuracy increase over the base model confirms the benefits of a personalised model when inferring stress and the ability to perform transfer learning on-device simplifies the process of developing a personalised model.

## 5 CONCLUSION AND FUTURE WORK

We have proposed a novel method for personalising an affective 1D CNN model on-device with transfer learning techniques. Edge computing interfaces have been used to collect a small amount of real-world personalised labelled data for 3 target subjects where transfer learning is then used to personalise a 1D CNN trained using data collected from a controlled stressor experiment. As only a small sample of labelled data is required it saves time, labour and money while also improving model personalisation and performing across different domains. Edge computing has enabled the transfer learning model to be trained and ran on-device allowing for the real-time inference of stress. The transfer learning approach improved model accuracy from 82.5% to an average of 93.9% for the three target subjects.

In future work, the inference from the personalised models developed will be evaluated against self-reports to confirm real-world accuracy. Additionally, it would be beneficial to explore other devices than the Raspberry Pi that may significantly reduce the time required to apply the transfer learning approach. Interventional feedback (Woodward et al., 2018) should also be used in future iterations of the tangible interfaces enabling feedback to be issued as soon as stress is inferred, improving wellbeing in real-time.